%% file: main.tex
\pdfoutput=1

\documentclass[11pt]{article}

\usepackage[]{EMNLP2023}

\usepackage{times}
\usepackage{latexsym}

\usepackage[T1]{fontenc}

\usepackage[utf8]{inputenc}

\usepackage{microtype}

\usepackage{inconsolata}
\usepackage{enumitem}
\usepackage{adjustbox}
\usepackage{multirow}
\usepackage{caption}
\usepackage{amsmath}
\usepackage{graphicx}
\usepackage[symbol]{footmisc}
\usepackage[justification=centering]{caption}

\usepackage{listings}
\usepackage{xcolor}

\definecolor{codegreen}{rgb}{0,0.6,0}
\definecolor{codegray}{rgb}{0.5,0.5,0.5}
\definecolor{codepurple}{rgb}{0.58,0,0.82}
\definecolor{backcolour}{rgb}{0.95,0.95,0.92}

\lstdefinestyle{mystyle}{
    backgroundcolor=\color{backcolour},   
    commentstyle=\color{codegreen},
    keywordstyle=\color{magenta},
    numberstyle=\tiny\color{codegray},
    stringstyle=\color{codepurple},
    basicstyle=\ttfamily\footnotesize,
    breakatwhitespace=false,         
    breaklines=true,                 
    captionpos=b,                    
    keepspaces=true,                 
    numbers=left,                    
    numbersep=5pt,                  
    showspaces=false,                
    showstringspaces=false,
    showtabs=false,                  
    tabsize=2
}

 \newcommand\blfootnote[1]{%
  \begingroup
  \renewcommand\thefootnote{}\footnote{#1}%
  \addtocounter{footnote}{-1}%
  \endgroup
}

\title{Sig-Networks Toolkit: Signature Networks for Longitudinal Language Modelling}

\author{
\textbf{Talia Tseriotou$^{1*}$, Ryan Sze-Yin Chan$^{2*}$, Adam Tsakalidis$^{1,2}$, Iman Munire Bilal$^{3}$,}\\
\textbf{Elena Kochkina$^{1\ddag}$, Terry Lyons$^{2,4}$, Maria Liakata$^{1,2,3}$}\\
       $^1$Queen Mary University of London, $^2$The Alan Turing Institute,\\
       $^3$University of Warwick, $^4$University of Oxford\\
      \tt  t.tseriotou@qmul.ac.uk; rchan@turing.ac.uk}

\begin{document}
\maketitle

\blfootnote{* Indicates equal contribution.}
\blfootnote{\ddag Work done while at Queen Mary University of London.}

\renewcommand*{\thefootnote}{\arabic{footnote}}

\input{0abstract.tex}

\input{1introduction.tex}

\input{2related.tex}

\input{3method}

\input{4system.tex}

\input{5experiments}

\input{6results}

\input{7conclusion}

\input{limitations.tex}

\section*{Ethics Statement}
The current project focuses on providing a toolkit for facilitating research and applications in longitudinal modelling. This is showcased in three tasks, two of which employ existing datasets (TalkLife and AnnoMI) and one is a re-interpretation of an existing public dataset (LRS).

Since the TalkLife dataset involves sensitive user generated social media content, Ethics approval was received from the Institutional Review Board (IRB) of the corresponding ethics board
of the University of Warwick prior to engaging in longitudinal modelling with this dataset. Thorough data analysis, data sharing policies to protect sensitive information and data anonymisation were used to address ethical considerations around the nature of such data  \citep{mao2011loose, kekulluoglu2020analysing}. Access to TalkLife's data was obtained through the submission of a project proposal and the approval of the corresponding license by TalkLife\footnote{\url{https://www.talklife.com/}}. TalkLife data were maintained and experiments were ran through a secure server accessible only by our group members. While we release code examples and results, we do not release any data, labels, models or preprocessing associated with TalkLife data in our git repository.

The AnnoMI dataset is publicly available and is based on transcribed videos of therapy sessions which are enacted.

The LRS dataset is a re-interpretation of the RumourEval 2017 dataset to reflect switches in stance over time. RumourEval-2017 is a well established dataset for stance and rumour verification. The longitudinal stance extension of the dataset allows studying the changes in public stance over time.

Developing methods for longitudinal modeling is an important research direction for better interpretation of events. Potential risks from the application of our work in being
able to identify moments of change in individuals’
timelines are akin to those in earlier work on personal event identification from social media and
the detection of suicidal ideation. Potential mitigation strategies include restricting access to the code
base trained on TalkLife and annotation labels used for evaluation.

\section*{Acknowledgements}

This work was supported by a UKRI/EPSRC
Turing AI Fellowship to Maria Liakata (grant
EP/V030302/1), the Alan Turing Institute (grant
EP/N510129/1), Baskerville (a national accelerated compute resource under the EPSRC Grant EP/T022221/1), a DeepMind PhD Scholarship, an EPSRC (grant EP/S026347/1), the Data Centric Engineering Programme (under the Lloyd’s Register Foundation grant G0095), the Defence and Security Programme (funded by the UK Government), the Office for National Statistics \& The Alan Turing Institute (strategic partnership) and by the Hong Kong Innovation and Technology Commission (InnoHK Project CIMDA).

The authors would like to thank Kasra Hosseini and Nathan Simpson for their early contributions to the \texttt{nlpsig} library as well as Federico Nanni and the anonymous reviewers for their valuable feedback. 

\input{ref.tex}

\appendix

\input{appendix.tex}

\end{document}

%% file: 0abstract.tex
\begin{abstract}

We present an open-source, pip installable toolkit, \textbf{Sig-Networks}, the first of its kind for longitudinal language modelling. A central focus is the incorporation of Signature-based Neural Network models, which have recently shown success in temporal tasks. 
We apply and extend published research providing a full suite of signature-based models. Their components can be used as PyTorch building blocks in future architectures. Sig-Networks enables task-agnostic dataset plug-in, seamless pre-processing for sequential data, parameter flexibility, automated tuning across a range of models. 
We examine signature networks under three different NLP tasks of varying temporal granularity: counselling conversations, rumour stance switch and mood changes in social media threads, showing SOTA performance in all three, and provide guidance for future tasks. 
We release the Toolkit as a PyTorch package\footnote{\url{https://pypi.org/project/sig-networks/}} with an introductory video \footnote{\url{http://youtu.be/lrjkdfYf8Lo}}, Git repositories for preprocessing\footnote{\url{https://github.com/datasig-ac-uk/nlpsig/}} and modelling\footnote{\url{https://github.com/ttseriotou/sig-networks/}} including sample notebooks on the modeled NLP tasks.

\end{abstract}

%% file: 1introduction.tex
\section{Introduction}


Existing work on temporal and longitudinal modelling has largely focused on models that are task-oriented, including tracking mood changes in users' linguistic content \citep{tsakalidis2022identifying, tsakalidis2022overview}, temporal clinical document classification \citep{ng2023modelling}, suicidal ideation detection on social media \citep{cao2019latent, sawhney2021suicide}, real-time rumour detection \citep{liu2015real,kochkina2023evaluating}. 
Transformer-based models struggle to outperform more traditional RNNs in such tasks, highlighting their limitations in temporal settings~\citep{mullenbach2018explainable, yuan2022code}.
Inspired by the success of models with short- and long-term processing capabilities \citep{, didolkar2022temporal,tseriotou2023sequential} in producing compressed 
temporal representations, we develop a toolkit that applies Signature Network models \citep{tseriotou2023sequential} to various longitudinal tasks. Path signatures are capable of efficient and compressed encoding of sequential data, sequential pooling in
neural models, enhancement of short-term dependencies in linguistic timelines and encoding agnostic to task and time irregularities. 

\noindent We make the following contributions:
\begin{itemize}[noitemsep, topsep=0pt, leftmargin=*]
    \item We release an open-source pip installable 
    toolkit for longitudinal NLP tasks, \textbf{Sig-Networks}, including examples on several tasks to facilitate usability and reproducibility. 
    
    \item For data preprocessing for the Signature Networks models \citep{tseriotou2023sequential}, we introduce another pip installable library \texttt{nlpsig} which receives as input streams of textual data and returns streams of embeddings which can be fed into the models we discuss in this paper.
    
    \item We showcase SOTA performance on three longitudinal tasks with different levels of temporal granularity, including a new task and dataset -- longitudinal rumour stance, based on rumour stance classification \cite{zubiaga2016analysing,kochkina2018all}. We highlight best practices for adaptation to new tasks.
    
    \item Our toolkit allows for flexible adaptation to new datasets, preprocessing steps, hyperparameter choices, external feature selection and benchmarking across several baselines. 
    We provide the option of flexible building blocks such as Signature Window Network Units~\citep{tseriotou2023sequential} and their extensions,
    which can be used as a layer integrated in a new PyTorch model 
     or as a stand-alone model for sequential NLP tasks. We share NLP-based examples via notebooks, where users can easily plug in their own datasets. 
    
\end{itemize}

%% file: 2related.tex
\section{Related Work}

\textbf{Longitudinal NLP modelling} has been sporadically explored in tasks like semantic change detection \citep{bamler2017dynamic, yao2018dynamic, tsakalidis2020sequential, montariol2021scalable, rosin2022temporal} or dynamic topic modelling  \citep{he2014dynamic,gou2018constructing, dieng2019dynamic, grootendorst2022bertopic}. Such approaches have limited generalisability as they track the evolution of specific topics over long-periods of time. 
Social media data have given rise to longitudinal tasks such as mental health monitoring~\citep{sawhney2021suicide, tsakalidis2022overview}, stance detection and rumour verification \citep{kochkina2018all, chen2018call, kumar2019tree} requiring more fine-grained temporal modelling . Other tasks, like healthcare patient notes \citep{ng2023modelling} and dialogue act classification \citep{liu2017using, he2021speaker} are also longitudinal in nature.\\

\noindent\textbf{Path Signature}\label{sec:related:signatures} 
\citep{chen1958integration,lyons1998differential} 
  is a collection of iterated integrals studied in the context of solving differential equations driven by irregular signals. It provides a summary of complex un-parameterised data streams through an infinite graded sequence of important statistics. Thus, it produces a collection of statistics efficiently summarising important information about the path. 
Signatures are deemed invaluable in machine learning \cite{levin2013learning} as sequential feature transformers \citep{yang2016rotation, xie2017learning,yang2017developing,lyons2014feature,perez2018signature,morrill2020utilization}, 
or integrated components of neural models \citep{bonnier2019deep,liao2021logsig,tseriotou2023sequential}. 
However, they have only been sparsely explored within NLP \citep{wang2019path,wang2021modelling,biyong2020information}, addressing only sequentiality or temporality. Motivated by the wide range of longitudinal NLP tasks 
and the work by \citet{tseriotou2023sequential} we present a toolkit for neural sequential path signatures models achieving SOTA performance in a range of such tasks.\\

\noindent\textbf{Libraries for computing path signatures}\label{sec:related:signature_libraries}
include \texttt{roughpy}
, \texttt{esig}
, \texttt{iisignature} 
\citep{iisignature}, \texttt{signatory} 
\citep{kidger2021signatory} and \texttt{signax} (see links in Appendix \ref{app:siglibraries}). 
Currently, only \texttt{signatory} and \texttt{signax} offer differentiable computations of the signature and log-signature transforms on GPU (with PyTorch \citep{paszke2019pytorch} and JAX \citep{jax2018github}, respectively).

\noindent While the above libraries only perform signature computations, with \texttt{signatory} additionally allowing for data stream augmentation through convolutional neural networks, the Sig-Networks library provides users with a complete pipeline for the application of signature-based (Signature Network) models in longitudinal NLP tasks. 
In particular, Sig-Networks is a pip installable PyTorch library using \texttt{signatory} for differentiable computations of signature transforms on GPU, providing a range of off-the shelf models for task-agnostic longitudinal modeling.
Furthermore, the pip installable \texttt{nlpsig} library simplifies the data preprocessing for Signature Network models, by forming streams of embeddings to be directly fed into the models.

%% file: 3method.tex
\section{Methodological Foundations}\label{sec:method}

\begin{figure*}[h]
\centering
\captionsetup{justification=justified}
\includegraphics[width=.93\linewidth]{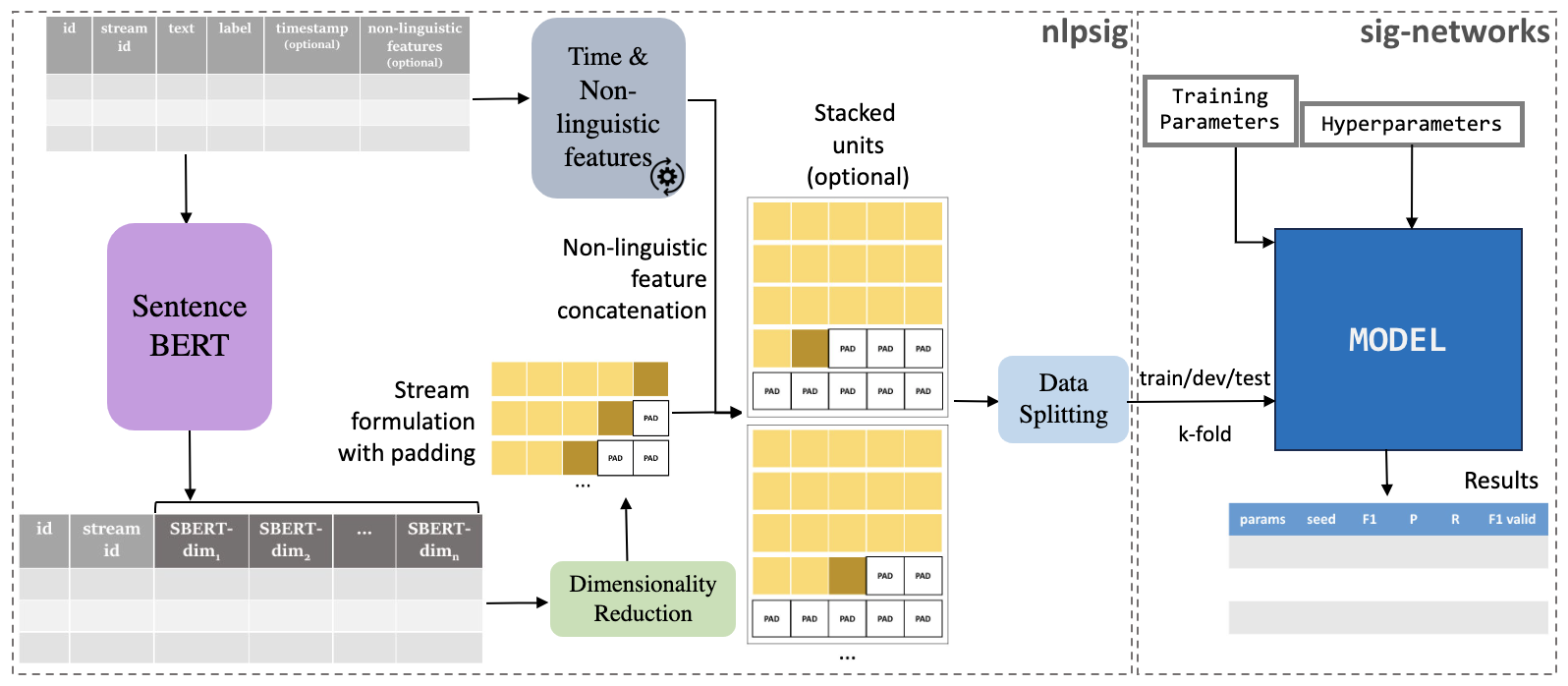}
\caption{Sig-Networks Tooklit Overview. \texttt{nlpsig} library (left side) obtains the input text, label and stream id per data point. The package allows for embedding extraction (i.e. SBERT) and its dimensionality reduction, with optional non-linguistic-feature processing and concatenation. For each data point a stream/window (padded if necessary) is formed including its ordered history. These are shifted and stacked for unit-based models. Data splitting with k-fold option is performed. \texttt{sig-networks} library (right side) enables PyTorch implementation for all Sig-Networks family and baseline models with user-specified training and hyper parameter inputs.} 
\label{fig:toolkitoverview}
\end{figure*}

\subsection{Task Formulation and Background}

\textbf{Longitudinal Task Formulation}. We use the following terminology throughout the paper:

\begin{itemize}[noitemsep, topsep=0pt, leftmargin=*]
    \item \textit{Data Point}: $d_{i}$, is a single piece of information at a given time
    , i.e. a post, tweet or utterance. 
    \item \textit{Data Stream}: $S^{[t_1,t_m]}$, is a series of chronologically ordered data points $\{d_{1}, \ldots, d_{m}\}$ at times $\{t_1, \ldots , t_m \}$, i.e. a timeline or a conversation.
\end{itemize}

\noindent For each $d_{i}$, we consider its historical data stream. We divide our models in two categories: (a) \textit{window-} and (b) \textit{unit-based}. In (a) we assume a window of  $|w|$ most recent historical data points of $d_{i}$, $H_i$=$\{d_{i-(w-1)}, \ldots, d_{i}\}$, as our modeling sequence. In (b), we follow  \citet{tseriotou2023sequential} to construct $n$  history windows, each of length $|w|$, shifted by $k$ points.\footnote{The total number of modeled data points is $k*n + (w-k)$.} The modeling sequence is given by $H_i = \{h_{i_{1}},$ ...,$ h_{i_{n-1}} ,h_{i_{n}}\}$ with the $q$th unit (of $w$ posts) defined as  $h_{i_{q}}$= $\{p_{i-(n-q)k-(w-1)},p_{i-(n-q)k-(w-2)}, ... ,p_{i-(n-q)k} \}. $\\

\noindent\textbf{Path Signatures Preliminaries}. 
In our formulation, the textual data stream is the equivalent of the path $P$ over an interval $[t_1,t_m]$ and the signature $S(P)$ is a pooling layer providing a transformed representation for these sequential data.
The signature is a collection of all $r$ iterated integrals along dimensions $c$: $S(P)_{t_1,t_m} = (1, S(P)^{1}_{t_1,t_m},...,S(P)^{c}_{t_1,t_m}, S(P)^{1,1}_{t_1,t_m},S(P)^{1,2}_{t_1,t_m},$ ... , $S(P)^{c,c}_{t_1,t_m}, ..., S(P)^{i_1, i_2, \cdots, i_r}_{t_1,t_m}, ...)$.
Since the iterated integrals can go up to infinite dimensions, a degree of truncation $N$ (i.e. up to N-folded integrals) is commonly used. A higher $N$ leads to a larger feature space
. \textit{Log}-signatures' 
output feature space increases less rapidly with input dimensions $c$, and depth $N$, allowing a more compressed representation. Sig-Networks allows for the selection of the desired $N$ and the implementation of signatures or log-signatures. We use $N$=3 and log-signatures which achieved the best performance.

\subsection{System Overview}

Fig.~\ref{fig:toolkitoverview} shows the overview of our Sig-Networks toolkit. The system receives a task-agnostic dataset of linguistic data streams. These can optionally include a set of pre-computed linguistic embeddings for each data point (e.g. post), timestamps and non-linguistic external features. Linguistic embeddings can also be computed by the system and then dimensionally reduced using a selected method~(\S\ref{sec:method:feature_encoding}). Timestamps can be processed to produce and normalise time-related features. The data points are then chronologically ordered and padded based on either a \textit{window-} or a \textit{unit-basis}. Data splitting for model training is performed by the relevant module (\S\ref{sec:datamodules}), providing a range of options (including k-fold,  stratification, user-defined splits). A range of baseline and Signature Network models are available for training (\S\ref{sec:method:sw_models},\ref{sec:system:training},\ref{sec:system:model}) through user defined parameters, integrating hyperparameter tuning functions for task-based optimal parameter selection. 

\begin{figure}[t]
\includegraphics[width=\linewidth]{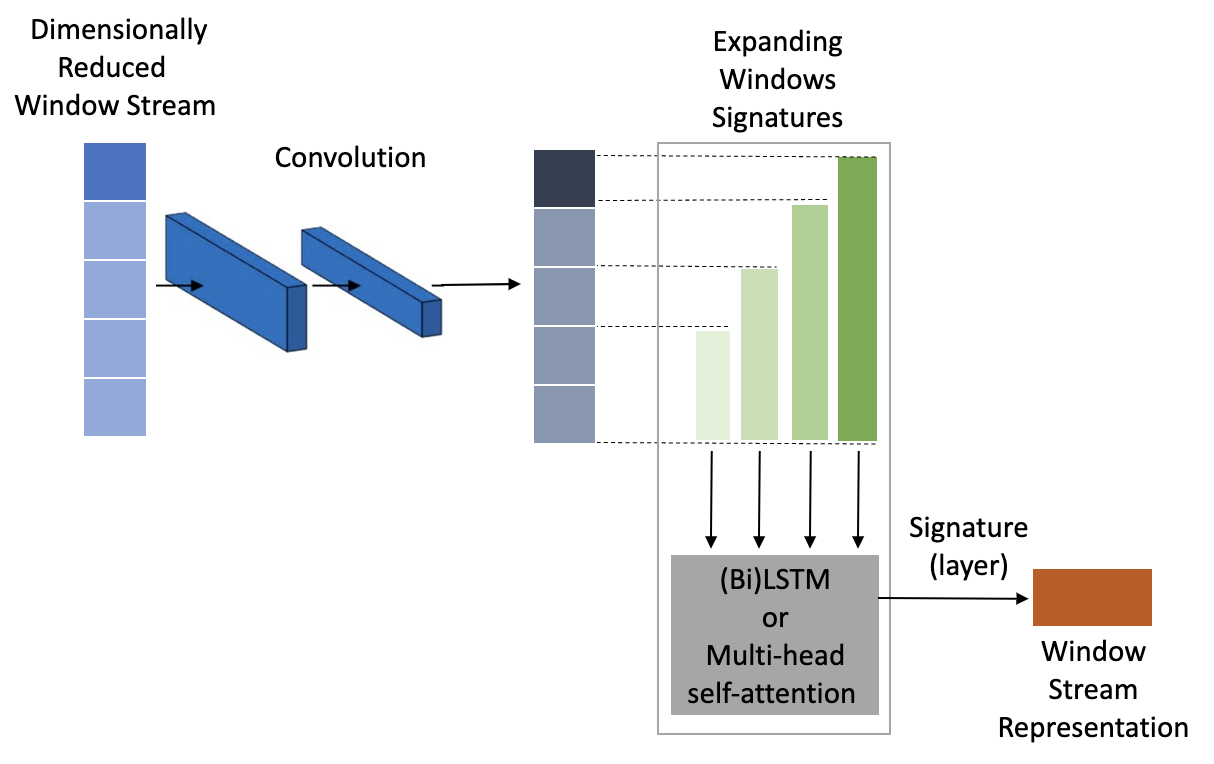}
\caption{Signature Window Unit and its variations.}
\label{fig:swunit}
\end{figure}

\subsection{Feature Encoding} \label{sec:method:feature_encoding}

Each data point is encoded in a high-dimensional space using SentenceBERT (SBERT) \citep{reimers2019sentence} to derive semantically meaningful embeddings. 
Our toolkit provides different sentence encoding options (\S\ref{sec:datamodules})~\footnote{We recommend 384-dim embeddings to facilitate dimensionality reduction required for input to signature transforms.} 
and multiple options for dimensionality reduction (\S\ref{sec:datamodules}). We found UMAP to perform slightly better. Sig-Networks also caters for time-related and external feature incorporation. 
On the time-related feature front, the toolkit provides a range of timestamp-derived features and normalisation methods, which account for temporality in the task according to its characteristics as well as for improved performance (\S\ref{sec:datamodules} \& Appendix~\ref{app:guidance}). 
 External information and domain-specific features can be either included as part of the stream feature space, $c$, or concatenated at the output of the model.

\subsection{Signature Network Models} \label{sec:method:sw_models}

The Signature Network model family forms an extension of the work by \citet{tseriotou2023sequential} on combining  signatures with neural networks for longitudinal language modeling. 
We present a range of models (\S\ref{sec:sw_models}) based on the foundational Signature Window Network Unit (SWNU), which models the granular linguistic progression in a stream: it reduces a short input stream via a conv-1d layer operation, applies an LSTM on signatures on locally expanding windows of the stream and produces a stream representation via a signature pooling layer.

SWNU implementation is flexible, allowing selection between LSTM \textit{vs} BiLSTM, convolution-1d layer \textit{vs} convolution neural network (CNN), and the option to stack multiple such units to form a deeper network. Importantly, we also introduce a variant of SWNU (`SW-Attn'), replacing LSTM with a Multi-head self-attention with an add \& norm operation and a linear layer (Fig.~\ref{fig:swunit}). 

\begin{figure}[h]
\includegraphics[width=.84\linewidth]{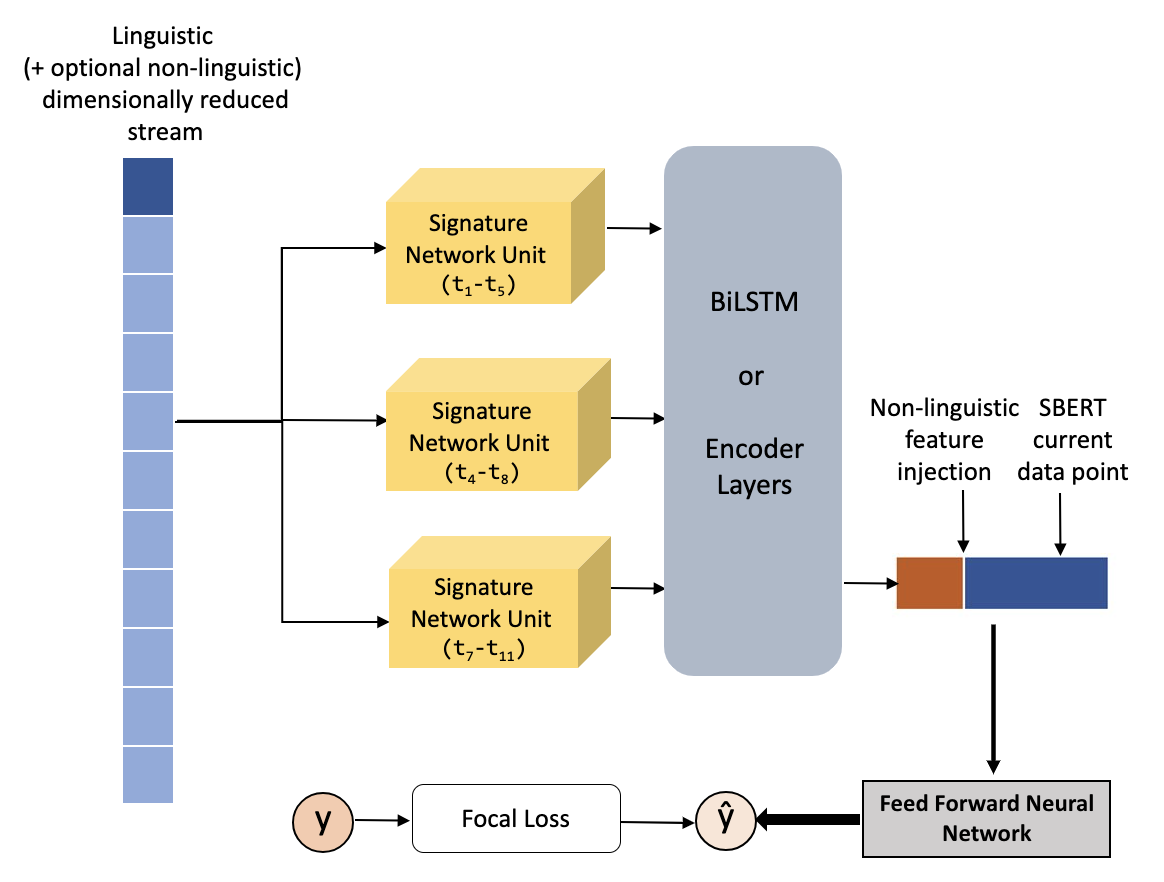}
\caption{Seq-Sig-Net and its variations using SWNU (yellow, see Fig.~\ref{fig:swunit}) on a sample length of 11 points.}
\label{fig:seqsignet}
\end{figure}

Furthermore, the toolkit allows for the flexible use of Seq-Sig-Net (the best performing model by \citet{tseriotou2023sequential}), which sequentially models SWNU units through a BiLSTM, preserving the local sequential information and capturing long-term dependencies. Further available variants of Seq-Sig-Net include SW-Attn+BiLSTM (replacing SWNU with a SW-Attn unit) and SW-Attn+Encoder (replacing BiLSTM with stacked Encoder layers on top of learnable unit embeddings). 
The final representation is pooled through a trainable [CLS] token. The number of stacked layers is user defined (see Fig.~\ref{fig:seqsignet}). For all Sig-Network models, we follow the same formulation as \citet{tseriotou2023sequential}, by concatenating the SBERT vectors of the current data point with the learnable stream representation and passing it through a feedforward network for classification using focal loss \citep{lin2017focal}. The system provides flexibility with respect to the number of hidden layers and the optional addition of external features. It also provides separate classes for the signature units so they can be incorporated in new architectures.

%% file: 4system.tex
\section{System Components} 

As shown in Fig.~\ref{fig:toolkitoverview}, the toolkit is split up into two pip installable Python libraries. a) \textbf{\texttt{nlpsig}}: SBERT vector extraction, data pre-processing including dimensionality reduction of SBERT streams and construction of model inputs and b) \textbf{\texttt{sig-networks}}: PyTorch implementations of our models and functions for model training/evaluation. 

\subsection{Data Preparation Modules in \texttt{nlpsig}} \label{sec:datamodules}

These modules perform data loading and preprocessing. The users can load their 
temporally sorted dataset, with a minimum of a \textit{stream-id} (identifying the stream that a data point belongs to), \textit{text} and \textit{label} columns. 
\texttt{nlpsig} allows for loading pre-computed embeddings for the data points or calculating them using any pretrained or custom model from the \texttt{sentence-transformer} and \texttt{transformer} libraries via the \texttt{nlpsig.encode\_text} modules.\footnote{ \href{https://nlpsig.readthedocs.io/en/latest/encode\_text.html}{encode\_text}}

Utilising signatures typically requires dimensionality reduction of the data point embeddings (\S\ref{sec:method:feature_encoding}). \texttt{nlpsig} provides several options via the \texttt{nlpsig.DimReduce}\footnote{ \href{https://nlpsig.readthedocs.io/en/latest/dimensionality\_reduction.html}{dimensionality\_reduction}} class: UMAP \cite{mcinnes2018umap}, Gaussian Random Projections \citep{bingham2001random,achlioptas2003database}, PPA-PCA \cite{mu2018all}, PPA-PCA-PPA \citep{raunak2019effective}. The \texttt{nlpsig.PrepareData}\footnote{\href{https://nlpsig.readthedocs.io/en/latest/data\_preparation.html}{data\_preparation}} class is used to process the data and obtain streams of dimension-reduced embeddings as input to the Signature Network family of models (see \S\ref{sec:method:sw_models}).

If the dataset includes timestamps, we automatically compute several time-derived variables 
with different standardisation options. These variables include but are not limited to chronologically ordered stream indices, time difference between consecutive data points and date as fraction of the year. External non-linguistic features can also be included in the dataset and model. The toolkit provides the flexibility of including these features as part of the path stream and/or concatenated in the output with the SBERT representation of the current data point (see Appendix \ref{app:guidance}). 
There are wrapper functions in the \texttt{sig-networks} package~(\texttt{sig\_networks.obtain\_SWNU\_input}, \texttt{sig\_networks.obtain\_SeqSigNet\_input}) to easily obtain the padded input for each model. Since the \texttt{nlpsig} library allows for more flexibility in constructing streams of embeddings, customisation of these wrapper functions is encouraged for different datasets or tasks.

\subsection{Training}\label{sec:system:training}
Through \texttt{nlpsig.classification\_utils}, the toolkit allows for k-fold cross validation or a single train/test split.\footnote{\href{https://nlpsig.readthedocs.io/en/latest/classification\_utils.html}{classification\_utils}} Splits can be completely random,  stratified (for streams via \texttt{split\_ids}), or pre-defined (via \texttt{split\_indices}). If a subset of the dataset is leveraged for classification (e.g. single-speaker classification in dialogue), the user can define such indices in \texttt{path\_indices}. 
For training, the user can select the loss function (cross-entropy, focal loss), a validation metric and specify the early stopping patience. Off the shelf hyperparameter tuning functions are available via grid search.

\noindent

\subsection{Model Modules} \label{sec:system:model}

Model modules allow for the flexible training of each model. 
PyTorch classes for the building blocks of our models are provided separately to encourage their novel integration in other systems (e.g. see Appendix~\ref{app:model_modules}). 
The toolkit can be used to benchmark datasets using: BERT, feedforward network with(out) historical stream information and BiLSTM. For Sig-Network family models, we provide options for choosing: 1. $N$, truncation degree, 2. signatures or log-signatures, 3. pooling options in the units, 4. LSTM or BiLSTM in SWNU, 5. dimensionality reduction of Conv-1d or CNN and their dimensions in the unit, 6. combination method of historical signature modelled stream with current SBERT data point and external features, 7. number of encoder layers, 8. path chronological reversion. Importantly, the user can assess their task of interest and define the window size $w$, number of units $n$, and shift $k$ (\S\ref{sec:performance}). After model tuning one can access the trained model object, a set of results for all seeds and hyperparameters, and a set of results for the best hyperparameters.

%% file: 5experiments.tex
\section{Experiments} \label{sec:experiments}

\subsection{Tasks and Datasets} \label{sec:tasks_n_data}

We demonstrate the applicability of Sig-Networks across three 
longitudinal sequential classification tasks of different temporal granularity. For all tasks we consider the current data point, its timestamp and its historical stream. 

\noindent\textbf{Moments of Change (MoC)}.
 Given sequences of users' posts,
MoC identification involves the assessment of a user's self-disclosed mood conveyed in each post with respect to the user's recent history as one of 3 classes: \textit{Switches} (IS): sudden mood shift from positive to negative, or vice versa; \textit{Escalations} (IE): gradual mood progression from neutral/positive to more positive, or from neutral/negative to more negative; or \textit{None} (O): no change in mood \citep{tsakalidis2022identifying}.
The dataset is \textbf{\emph{TalkLife MoC}}: 18,702 posts (500 user timelines; 1-124 posts each). Annotation was performed on the post-level with access to the entire timeline.

\noindent\textbf{Counselling Dialogue Classification}. \label{sec:taskdefannomi}
Given the data stream of utterances during a counselling dialogue between a therapist and a client, the task is to categorise client's utterances into one of 3 classes: 
\textit{Change}: client seems convinced towards positive behaviour change; \textit{Sustain}: client shows resistance to change; \textit{Neutral}: client shows neither leaning nor resistance towards change. We utilise therapist and client utterances in the stream, while classifying only client utterances. The dataset is \textbf{\emph{Anno-MI}}~\citep{wu2022anno}: 133 motivational interviews (MI), 9,699 utterances (4,817 client utterances), sourced from effective and ineffective MI videos on YouTube \& Vimeo. The videos were professionally transcribed and annotated by MI practitioners. 

\noindent\textbf{Stance Switch Detection}. 
The Stance Switch Detection task tracks the ratio of support/opposition towards the topic of a conversation at each point in time and captures switches in overall stance.  
This is a binary classification of each post in a conversation stream into: \textit{Switch}: switch  between the total number of oppositions (querying or denying) and supports or vice versa; or \textit{No Switch}: either the absence of a switch or cases where the numbers of supporting and opposing posts are equal.
For this task we introduce a new dataset, \textbf{\emph{Longitudinal Rumour Stance (LRS)}}, a longitudinal version of the RumourEval-2017 dataset \citep{gorrell2019semeval}.
It consists of Twitter conversations around 
newsworthy events. The source tweet of the conversation conveys a rumourous claim, discussed by tweets in the stream. 
In 325 conversations 5,568 posts are labelled based on their stance towards the claim in the corresponding source tweet as either \textit{Supporting}, \textit{Denying}, \textit{Questioning} or \textit{Commenting}. 
We convert conversation structure and labels into a Longitudinal Stance Switch Detection task. 
Conversations are converted from tree-structured into linear timelines 
to obtain chronologically ordered lists. Then we convert the original stance labels into \textit{Switch} and \textit{No Switch} categories based on the numbers of supporting tweets versus denying and questioning ones at each point in time. 

\subsection{Models and Baselines}\label{sec:sw_models}
Using our toolkit, we perform 5-fold cross-validation, repeatedly with 3 seeds (see Appendix \ref{app:experiments} for full details) and compare against the following baselines:
\vspace{.2cm}

\noindent\textbf{BERT(focal/ce)}: data point-level (stream-agnostic) BERT \citep{devlin2018bert} fine-tuned using the alpha-weighted focal loss \citep{lin2017focal} or cross-entropy, respectively.


\noindent\textbf{FFN}: data point-level Feedforward Network (FFN)  operating on SBERT of the current point.

\noindent\textbf{FFN History}: stream-level FFN operating on the concatenated SBERT vectors of the current point and the average of its historical stream.

\noindent\textbf{BiLSTM} with a single layer operating on a specified number of historical data points.

\vspace{.25cm}
\noindent Our Sig-Networks Family Models are:

\noindent\textbf{SWNU} \citep{tseriotou2023sequential} uses expanding signature windows fed into an LSTM. We modify the unit to use a BiLSTM and improved padding.

\noindent\textbf{SW-Attn}: Same as SWNU but with Multi-head attention instead of an LSTM.

\noindent\textbf{Seq-Sig-Net}: Sequential Network of SWNU units using a BiLSTM as in \citet{tseriotou2023sequential}.

\noindent\textbf{SW-Attn+BiLSTM}: Seq-Sig-Net with SW-Attn unit instead of SWNU.

\noindent\textbf{SW-Attn+Encoder} SW-Attn+BiLSTM with two Encoder layers using unit-level learnable embeddings instead of the BiLSTM .

%% file: 6results.tex
\section{Results and Discussion} \label{sec:results}

\begin{table*}[ht]
\begin{adjustbox}{max width=\textwidth,center}
\begin{tabular}{|l|lll|lll|lll|}
\hline
\multicolumn{1}{|c|}{\multirow{2}{*}{\textbf{Model}}} & \multicolumn{3}{c|}{\textbf{Anno-MI}} & \multicolumn{3}{c|}{\textbf{LRS}} & \multicolumn{3}{c|}{\textbf{TalkLife}} \\ 
\multicolumn{1}{|c|}{} & \multicolumn{3}{c|}{\textbf{(3-class)}} & \multicolumn{3}{c|}{\textbf{ (2-class)}} & \multicolumn{3}{c|}{\textbf{(3-class)}} \\ \hline
BERT (focal) & \multicolumn{3}{c|}{.519} & \multicolumn{3}{c|}{.589} & \multicolumn{3}{c|}{.531} \\ \hline
BERT (ce) & \multicolumn{3}{c|}{.501} & \multicolumn{3}{c|}{.596} & \multicolumn{3}{c|}{.521} \\ \hline
FFN & \multicolumn{3}{c|}{.512} & \multicolumn{3}{c|}{.581} & \multicolumn{3}{c|}{.534} \\ \hline
FFN History & \multicolumn{3}{c|}{.520} & \multicolumn{3}{c|}{.625} & \multicolumn{3}{c|}{.537} \\ \hline
BiLSTM ($w=5$) & \multicolumn{3}{c|}{.517} & \multicolumn{3}{c|}{.637} & \multicolumn{3}{c|}{.544} \\ \hline
SWNU ($w=5$) & \multicolumn{3}{c|}{.522} & \multicolumn{3}{c|}{.670} & \multicolumn{3}{c|}{\textbf{.563}} \\ \hline
SW-Attn ($w=5$) & \multicolumn{3}{c|}{.515} & \multicolumn{3}{c|}{.667} & \multicolumn{3}{c|}{.556} \\ \hline
\textbf{History Length} & \multicolumn{1}{c|}{\textbf{11}} & \multicolumn{1}{c|}{\textbf{20}} & \multicolumn{1}{c|}{\textbf{35}} & \multicolumn{1}{c|}{\textbf{11}} & \multicolumn{1}{c|}{\textbf{20}} & \multicolumn{1}{c|}{\textbf{35}} & \multicolumn{1}{c|}{\textbf{11}} & \multicolumn{1}{c|}{\textbf{20}} & \multicolumn{1}{c|}{\textbf{35}} \\ 
\textbf{\#units ($w$=5, $k$=3)} & \multicolumn{1}{c|}{\textbf{3}} & \multicolumn{1}{c|}{\textbf{6}} & \multicolumn{1}{c|}{\textbf{11}} & \multicolumn{1}{c|}{\textbf{3}} & \multicolumn{1}{c|}{\textbf{6}} & \multicolumn{1}{c|}{\textbf{11}} & \multicolumn{1}{c|}{\textbf{3}} & \multicolumn{1}{c|}{\textbf{6}} & \multicolumn{1}{c|}{\textbf{11}} \\ \hline
BiLSTM & \multicolumn{1}{r|}{.518}  & \multicolumn{1}{r|}{.507} & \multicolumn{1}{r|}{.510} & \multicolumn{1}{r|}{.657} & \multicolumn{1}{r|}{.648} & \multicolumn{1}{r|}{.648} & \multicolumn{1}{r|}{.539}  & \multicolumn{1}{r|}{.533}& \multicolumn{1}{r|}{.525} \\ \hline
SWNU & \multicolumn{1}{r|}{.522} & \multicolumn{1}{r|}{.512} & \multicolumn{1}{r|}{.493} & \multicolumn{1}{r|}{.671} & \multicolumn{1}{r|}{.654} & \multicolumn{1}{r|}{\underline{.673}} & \multicolumn{1}{r|}{.550} & \multicolumn{1}{r|}{.537} & \multicolumn{1}{r|}{.539} \\ \hline
SW-Attn & \multicolumn{1}{r|}{.517} & \multicolumn{1}{r|}{.508} & \multicolumn{1}{r|}{.508} & \multicolumn{1}{r|}{.659} & \multicolumn{1}{r|}{.665} & \multicolumn{1}{r|}{.661} & \multicolumn{1}{r|}{.547} & \multicolumn{1}{r|}{.541} & \multicolumn{1}{r|}{.539} \\ \hline
Seq-Sig-Net & \multicolumn{1}{r|}{\textbf{.525}} & \multicolumn{1}{r|}{\underline{.523}} & \multicolumn{1}{r|}{.517} & \multicolumn{1}{r|}{.672} & \multicolumn{1}{r|}{\textbf{.678}} & \multicolumn{1}{r|}{.654} & \multicolumn{1}{r|}{\textbf{.563}} & \multicolumn{1}{r|}{\underline{.561}} & \multicolumn{1}{r|}{.559} \\ \hline
SW-Attn+BiLSTM & \multicolumn{1}{r|}{.511} & \multicolumn{1}{r|}{.514} & \multicolumn{1}{r|}{.515} & \multicolumn{1}{r|}{.663} & \multicolumn{1}{r|}{.657} & \multicolumn{1}{r|}{.660} & \multicolumn{1}{r|}{.554} & \multicolumn{1}{r|}{.557} & \multicolumn{1}{r|}{.550} \\ \hline
SW-Attn+Encoder & \multicolumn{1}{r|}{.498} & \multicolumn{1}{r|}{.506} & \multicolumn{1}{r|}{.505} & \multicolumn{1}{r|}{.664} & \multicolumn{1}{r|}{.657} & \multicolumn{1}{r|}{.662} & \multicolumn{1}{r|}{.552} & \multicolumn{1}{r|}{.552} & \multicolumn{1}{r|}{.545} \\ \hline
\end{tabular}
\end{adjustbox}
\caption{Results (macro-average F1) of the Sig-Networks toolkit models on our three tasks for different History Lengths. \textbf{Best} and \underline{second best} scores are highlighted.}
\label{tab:demoallresults}
\end{table*}

\subsection{Performance comparison} \label{sec:performance}

Signature Network models show top performance, with Seq-Sig-Net achieving SOTA or on-par performance with SWNU across all tasks (see Table \ref{tab:demoallresults}, detailed in Appendix \ref{app:detailedresults}). On LRS the best model is Seq-Sig-Net with window length $w$=20 posts (F1=.678), while on Anno-MI the best model is also Seq-Sig-Net but for $w$=11 (F1=.525). In TalkLife, Seq-Sig-Net and SWNU both reach top performance (F1=.563). The difference of optimal window length across tasks 
relates to the characteristics of each dataset (see Table \ref{tab:statistics} and next paragraph). Additionally, 
the performance of BiLSTM peaks within the same range of history length, different for each task, denoting the best performing models depend on the temporal granularity of the task. 
Lastly, SW-Attn+Encoder performs better than  SW-Attn+BiLSTM regardless of task and history length, further highlighting the importance of sequential modelling for these tasks.

Seq-Sig-Net outperforms all models across tasks in modeling long-term effects, making it particularly appealing for highly longitudinal tasks; SWNU has the best performance when modeling short linguistic streams. In LRS and TalkLife Sig-Networks outperforms all baselines, for each history length. For Anno-MI, the least longitudinal task due to the short mean/median consecutive sequences of Change/Sustain utterances (see Table \ref{tab:statistics}), we conjecture that most of the performance gain in including historical dialogue information is due to adding more context rather than sequential modelling. This is apparent from the small performance gains of Seq-Sig-Net models compared to FFN History and BERT (focal) 
versus the much starker performance improvement in the other tasks.



\subsection{Time-scale analysis} \label{sec:datasetchar}

The degree of temporal granularity across datasets ranges from seconds in Anno-MI, minutes in LRS and hours in TalkLife (Table \ref{tab:statistics}), showing the generalisability of Signature-Networks. 
TalkLife has an average of 1.58/4.12 consecutive Switches/Escalations and a similar average of such events (1.77/4.03 respectively) in each data stream, meaning that the task benefits from good granularity on short modeling windows. 
This can be provided by both SWNU (window of 5 posts) and a short Seq-Sig-Net of 3 units. Anno-MI presents even shorter sequences of consecutive Change/Sustain intentions (2.21/1.68), but the average number of such events in each conversation is higher (8.86/4.05), therefore benefiting from being less sequential in terms of short-term dependencies but being more sequence dependent on series of windows. Finally, LRS is the most longitudinal task in our experiments showing the highest mean number of consecutive switches (8.52), therefore benefiting from more units in Seq-Sig-Net.

\begin{table}[ht]
\begin{adjustbox}{max width=\linewidth}
\begin{tabular}{|l|ll|l|ll|}
\hline
\multicolumn{1}{|c|}{\multirow{3}{*}{\textbf{Dataset}}} & \multicolumn{2}{c|}{\textbf{Anno-MI}}                                                            & \multicolumn{1}{c|}{\textbf{Longitudinal}}         & \multicolumn{2}{c|}{\textbf{TalkLife}}                                                          \\ 
\multicolumn{1}{|c|}{} & \multicolumn{2}{c|}{}                                                            & \multicolumn{1}{c|}{\textbf{Rumour Stance}}         & \multicolumn{2}{c|}{\textbf{MoC}}                                                          \\ \cline{2-6} 
\multicolumn{1}{|c|}{}                                  &  \multicolumn{1}{c|}{\textbf{Change}} & \multicolumn{1}{c|}{\textbf{Sustain}} & \multicolumn{1}{c|}{\textbf{Switch}} & \multicolumn{1}{c|}{\textbf{   Switch    }} & \multicolumn{1}{c|}{\textbf{Escalation}} \\ \hline
{Mean Point Time Diff.}             & \multicolumn{2}{c|}{5sec}                                                               & \multicolumn{1}{c|}{1hr 26min 40sec}             & \multicolumn{2}{c|}{6hr 51min 11sec}                                                 \\ \hline
{Median Point Time Diff.}           & \multicolumn{2}{c|}{3sec}                                                               & \multicolumn{1}{c|}{1min 39sec}                     & \multicolumn{2}{c|}{59min 38sec}                                                          \\ \hline
{Mean consecutive events}                     & \multicolumn{1}{r|}{2.21}            & \multicolumn{1}{r|}{1.68}             & \multicolumn{1}{r|}{8.52}            & \multicolumn{1}{r|}{1.58}            & \multicolumn{1}{r|}{4.12}                  \\ \hline
{Median consecutive events}                   & \multicolumn{1}{r|}{1}               & \multicolumn{1}{r|}{1}                               & \multicolumn{1}{r|}{4}               & \multicolumn{1}{r|}{1}               & \multicolumn{1}{r|}{3}                  \\ \hline
{Mean no. of events in stream}              & \multicolumn{1}{r|}{8.86}            & \multicolumn{1}{r|}{4.05}             &  \multicolumn{1}{r|}{6.45}            & \multicolumn{1}{r|}{1.77}            & \multicolumn{1}{r|}{4.03}                      \\ \hline
{Median no. of events in stream}            &  \multicolumn{1}{r|}{5}               & \multicolumn{1}{r|}{3}                                  & \multicolumn{1}{r|}{0}               & \multicolumn{1}{r|}{1}               & \multicolumn{1}{r|}{1}                  \\ \hline
\end{tabular}
\end{adjustbox}
\caption{Dataset Statistics on time and event length.}
\label{tab:statistics}
\end{table}

%% file: 7conclusion.tex
\section{Conclusion}
We present the Sig-Networks toolkit, which allows for flexible modeling of longitudinal NLP classification tasks using Signature-based Network models 
\citep{tseriotou2023sequential}, proposing improvements and variants. We test our system on three NLP classification tasks of different domains and temporal granularity and show SOTA performance against competitive baselines, while also shedding light into temporal characteristics which affect optimal model selection. The toolkit is made available as a PyTorch package with examples, making it easy to plug-in new datasets for future model extensions. 

In the future we are planning to
provide further flexibility in the toolkit, by enabling the integration of signature libraries beyond \texttt{signatory} for signature computations. This will facilitate its extension to deep learning frameworks beyond PyTorch. Additionally, we would like to allow for the selection of a non-linguistic feature subset to serve as part of the stream and of a different subset to be concatenated with the SBERT representation, rather than having a single processing option for the full set of such features. We also aim to enrich the examples corpus of our repository with additional longitudinal NLP tasks and collaborate with independent contributors on the integration of newly developed signature-based models.


%% file: limitations.tex
\section*{Limitations}

While the Sig-Networks library provides sequential models with very competitive performance on longitudinal NLP tasks, it comes with limitations. Firstly, it requires basic knowledge of Python, since it is available as a PyTorch library, and assumes integration in PyTorch systems. In the future, Additionally, its use on classification tasks requires labeled data, which can be expensive to obtain for tasks that require expert annotation. Although our tasks under examination are in English, we believe that this work is extensible to other languages. Since one of the initial steps for obtaining linguistic representations involves the use of a pretrained language model, we expect lower quality for low-resource languages where such pretrained models have poor performance or are non-existent.

Hyperparameter tuning including time feature selection, given that the timestamps are available, is often key in achieving competitive classification performance. We provide guidelines and expect the users to perform a thorough grid search if needed to reach a competitive performance. 
Lastly, we understand that our data point-level evaluation, which assesses predictions at each point in the stream in silo, can be lacking pattern identification on a stream level. We plan to address stream-level evaluation using the settings from \citet{tsakalidis2022identifying} in future work and we encourage users to cross-check performance with stream-level metrics.

%% file: ref.tex

\bibliographystyle{acl_natbib}

%% file: appendix.tex
\section{Experiment setup details} \label{app:experiments}

We train all models using PyTorch \citep{paszke2019pytorch} and Huggingface Transformers \citep{wolf2020transformers} for BERT, using the alpha-weighted focal loss \citep{lin2017focal}, except for BERT (ce). 

\noindent\textbf{SBERT representations}: As noted in \S\ref{sec:method:feature_encoding}, we use SentenceBERT (SBERT) \citep{reimers2019sentence} to encode each data point to obtain semantically meaningful embeddings. To do this with our toolkit, we used the \texttt{nlpsig.SentenceEncoder}\footnote{\url{https://nlpsig.readthedocs.io/en/latest/encode_text.html}} class which uses the \texttt{sentence-transformers} library. For each dataset, we obtained 384-dimensional embeddings using the "all-MiniLM-L12-v2" model\footnote{\url{https://huggingface.co/sentence-transformers/all-MiniLM-L12-v2}}.

\noindent\textbf{Model experiment settings}: In each of our experiments in \S\ref{sec:experiments}, we select the best model for each of the 5 folds using the best validation F1 macro-average score on 100 epochs with early stopping (patience set to 3). For training, we use the Adam optimiser \citep{kingma2014adam} with weight decay of $0.0001$. For all models, we use the alpha-weighted focal loss \citep{lin2017focal} with setting $\gamma=2$ and alpha of $\sqrt{1/p_{t}}$ where $p_{t}$ is the probability of class $t$ in the training data. The exception is for the BERT (ce) baseline model where we used the cross-entropy loss. For BERT, we used batch size of 8 during training due to limited GPU resources available for training on the secure data environment which hosted the TalkLife dataset. For the other models, we used batch size of 64.

For the TalkLife MoC dataset, we use the same train/test splits as in \citet{tsakalidis2022overview, tsakalidis2022identifying, tseriotou2023sequential}. Furthermore, we average the F1 macro-average performance over three random seeds, $(1, 12, 123)$. For Anno-MI and Longitudinal Rumour Stance datasets, we created the five folds using the \texttt{nlpsig.Folds} class\footnote{\url{https://nlpsig.readthedocs.io/en/latest/classification_utils.html}} class (with \texttt{random\_state=0}). Each fold constructed was used as a test and the rest as the training and validation data. Validation sets were formed on $33\%$ of the train set. When creating the folds, we stratify using the \texttt{transcript\_id} for Anno-MI and the conversation ID for Rumours to ensure there was no contamination between streams.

For each model, we perform a grid search for hyperparameter selection based on the validation set performance comparing F1 macro-average. For signature window models, prior to hyperparameter search, we performed dimensionality reduction on the SBERT embeddings using UMAP \citep{mcinnes2018umap} with the \texttt{umap-learn} Python library. Using the \texttt{UMAP}\footnote{\url{https://umap-learn.readthedocs.io/en/latest/api.html}} class in the library, we kept all default parameters besides \texttt{n\_neighbors=50}, \texttt{min\_dist=0.99} and \texttt{metric="cosine"}. In each of the signature window models, we reduced the SBERT embeddings to $15$ dimensions. For all models considered, the dropout rate was set to $0.1$.

In the rest of this section, we state the hyperparameters choices we had for each model. Note that the full results for each model that we trained (for each hyperparameter configuration and seed) as well as the best hyperparameters for each model and dataset can be found in the GitHub repository for the project in the \texttt{examples} folder\footnote{\url{https://github.com/ttseriotou/sig-networks/tree/main/examples}}.

\noindent\textbf{SWNU and Seq-Sig-Net}:
For the signature window networks which used the Signature Window Network Unit (SWNU) (\S\ref{sec:method:sw_models}, \ref{sec:sw_models}), hyperparameter selection was set through a grid search over the parameters: learning rate $\in [0.0005, 0.0003, 0.0001]$, LSTM hidden dimensions of SWNU $\in [10, 12]$, FFN hidden dimensions $\in [[32,32], [128,128], [512,512]]$ where $[h_{1},h_{2}]$ means a two hidden layer FFN of dimensions $h_{i}$ in the $i$th layer. For Seq-Sig-Net, the BiLSTM hidden dimensions $\in [300,400]$. We took the log-signature transform with depth (degree of truncation) 3. In each model run, the convolution-1d reduced dimensions is equal to the LSTM hidden dimensions (i.e. $10$ or $12$ here).

\noindent\textbf{SW-Attn and Seq-Sig-Net-Attention models}: 
For the signature window networks which used the Signature Window Attention Unit (\S\ref{sec:method:sw_models}, \ref{sec:sw_models}) hyperparameter selection was set through a grid search over the following parameters: learning rate $\in [0.0005, 0.0003, 0.0001]$, convolution-1d reduced dimensions $\in [10, 12]$, FFN hidden dimensions $\in [[32,32], [128,128], [512,512]]$. We took the log-signature transform with depth (degree of truncation) 3. While the toolkit allows you to easily stack multiple SW-Attn blocks, i.e. multiple iterations of taking the expanding window signatures and multi-head attention (with add+norm and a linear layer), we only have one block, $\texttt{num\_layer=1}$. 

For models using SW-Attn units, we must choose the number of attention heads to divide the resulting number of signature channels after taking streaming signatures. For models with conv-1d reduced dimensions set to $10$, \texttt{output\_channels=10}, we set \texttt{num\_heads=5} since after taking a log-signature of depth $3$, the output has dimension 385\footnote{\texttt{signatory.logsignature\_channels(10, 3)} can be used to compute this number.}. For models with \texttt{output\_channels=12}, we set \texttt{num\_heads=10} since the number of log-signature channels at depth $3$ for a path with $12$ channels is $650$.

\noindent\textbf{BERT}: We fine-tuned the \texttt{bert-base-uncased}\footnote{\url{https://huggingface.co/bert-base-uncased}} model on the Huggingface model hub, and used the \texttt{transformers} library and Trainer API for training the model. The only hyperparameter we performed a grid-search for was learning rate $\in [0.00005, 0.00001, 0.000001]$\footnote{Note in \texttt{transformers} (version 4.30.2), the default learning rate is $0.00005$}. For BERT, we found it was important to use a much lower learning rate than the ones we used for other models due to the larger number of parameters in the model.

\noindent\textbf{FFN models}: For models using a Feedforward Network (FFN), either operating on the SBERT embedding of the current point (\textbf{FFN}) or operating on a concatenation of the current SBERT embedding with the mean average of its historical stream (\textbf{FFN History}), we perform a hyperparameter search over learning rate $\in [0.001, 0.0005, 0.0001]$ and hidden dimensions $\in [[64,64], [128,128], [256,256], [512,512]]$.

\noindent\textbf{BiLSTM}: We apply a single layer BiLSTM on a specified number of historical SBERT embeddings for the data point. We perform a grid search over learning rate $\in [0.001, 0.0005, 0.0001]$ and hidden dimension sizes $[200, 300, 400]$.

\section{Results} \label{app:detailedresults}

We present class-level performance for each task in Tables \ref{tab:anno-mi}, \ref{tab:lrs} and \ref{tab:talklife}. 

\begin{table}[htb!]
\begin{adjustbox}{max width=\linewidth}\begin{tabular}{|l|lrrlrrlrrlrr|}
\hline
\multicolumn{1}{|c|}{\textbf{Model}}                                                                              & \multicolumn{3}{c|}{\textbf{Neutral(N)}}                                                               & \multicolumn{3}{c|}{\textbf{Change(C)}}                                                                        & \multicolumn{3}{c|}{\textbf{Sustain(S)}}                                                                    & \multicolumn{3}{c|}{\textbf{Macro-avg}}                                                                     \\ \hline
BERT (focal)                                                                                                      & \multicolumn{3}{c|}{.767}                                                                              & \multicolumn{3}{c|}{.449}                                                                                      & \multicolumn{3}{c|}{.339}                                                                                   & \multicolumn{3}{c|}{.519}                                                                                   \\ \hline
BERT (ce)                                                                                                         & \multicolumn{3}{c|}{\textbf{.784}}                                                                     & \multicolumn{3}{c|}{.442}                                                                                      & \multicolumn{3}{c|}{.277}                                                                                   & \multicolumn{3}{c|}{.501}                                                                                   \\ \hline
FFN                                                                                                               & \multicolumn{3}{c|}{.764}                                                                              & \multicolumn{3}{c|}{.424}                                                                                      & \multicolumn{3}{c|}{.347}                                                                                   & \multicolumn{3}{c|}{.512}                                                                                   \\ \hline
FFN History                                                                                                       & \multicolumn{3}{c|}{.761}                                                                              & \multicolumn{3}{c|}{.449}                                                                                      & \multicolumn{3}{c|}{.351}                                                                                   & \multicolumn{3}{c|}{.520}                                                                                   \\ \hline
BiLSTM (w=5)                                                                                                      & \multicolumn{3}{c|}{.753}                                                                              & \multicolumn{3}{c|}{.449}                                                                                      & \multicolumn{3}{c|}{.348}                                                                                   & \multicolumn{3}{c|}{.517}                                                                                   \\ \hline
SWNU (w=5)                                                                                                        & \multicolumn{3}{c|}{.762}                                                                              & \multicolumn{3}{c|}{.447}                                                                                      & \multicolumn{3}{c|}{.356}                                                                                   & \multicolumn{3}{c|}{.522}                                                                                   \\ \hline
SW-Attn (w=5)                                                                                                     & \multicolumn{3}{c|}{.749}                                                                              & \multicolumn{3}{c|}{{\underline{.450}}}                                                                                & \multicolumn{3}{c|}{.346}                                                                                   & \multicolumn{3}{c|}{.515}                                                                                   \\ \hline
\multicolumn{1}{|c|}{\multirow{2}{*}{\textbf{\begin{tabular}[c]{@{}c@{}}History Length \\ (units)\end{tabular}}}} & \multicolumn{4}{c|}{\textbf{11 (n=3)}}                                                                                                           & \multicolumn{4}{c|}{\textbf{20 (n=6)}}                                                                                                           & \multicolumn{4}{c|}{\textbf{35 (n=11)}}                                                                                                       \\ \cline{2-13} 
\multicolumn{1}{|c|}{}                                                                                            & \multicolumn{1}{c|}{\textbf{N}} & \multicolumn{1}{c|}{\textbf{C}} & \multicolumn{1}{c|}{\textbf{S}}    & \multicolumn{1}{c|}{\textbf{Macro-avg}} & \multicolumn{1}{c|}{\textbf{N}} & \multicolumn{1}{c|}{\textbf{C}}    & \multicolumn{1}{c|}{\textbf{S}} & \multicolumn{1}{c|}{\textbf{Macro-avg}} & \multicolumn{1}{c|}{\textbf{N}} & \multicolumn{1}{c|}{\textbf{C}} & \multicolumn{1}{c|}{\textbf{S}} & \multicolumn{1}{c|}{\textbf{Macro-avg}} \\ \hline
BiLSTM                                                                                                            & \multicolumn{1}{r|}{.746}       & \multicolumn{1}{r|}{.446}       & \multicolumn{1}{r|}{\textbf{.363}} & \multicolumn{1}{r|}{.518}               & \multicolumn{1}{r|}{.754}       & \multicolumn{1}{r|}{.446}          & \multicolumn{1}{r|}{.322}       & \multicolumn{1}{r|}{.507}               & \multicolumn{1}{r|}{.755}       & \multicolumn{1}{r|}{.446}       & \multicolumn{1}{r|}{.329}       & .510                                    \\ \hline
SWNU                                                                                                              & \multicolumn{1}{r|}{.761}       & \multicolumn{1}{r|}{.444}       & \multicolumn{1}{r|}{{\underline{.360}}}    & \multicolumn{1}{r|}{.522}               & \multicolumn{1}{r|}{.759}       & \multicolumn{1}{r|}{.440}          & \multicolumn{1}{r|}{.338}       & \multicolumn{1}{r|}{.512}               & \multicolumn{1}{r|}{.752}       & \multicolumn{1}{r|}{.413}       & \multicolumn{1}{r|}{.314}       & .493                                    \\ \hline
SW-Attn                                                                                                           & \multicolumn{1}{r|}{.759}       & \multicolumn{1}{r|}{{\underline{.450}}} & \multicolumn{1}{r|}{.341}          & \multicolumn{1}{r|}{.517}               & \multicolumn{1}{r|}{.754}       & \multicolumn{1}{r|}{.438}          & \multicolumn{1}{r|}{.333}       & \multicolumn{1}{r|}{.508}               & \multicolumn{1}{r|}{.749}       & \multicolumn{1}{r|}{.446}       & \multicolumn{1}{r|}{.330}       & .508                                    \\ \hline
Seq-Sig-Net                                                                                                       & \multicolumn{1}{r|}{{\underline{.769}}} & \multicolumn{1}{r|}{.446}       & \multicolumn{1}{r|}{.359}          & \multicolumn{1}{r|}{\textbf{.525}}      & \multicolumn{1}{r|}{{\underline{.769}}} & \multicolumn{1}{r|}{\textbf{.452}} & \multicolumn{1}{r|}{.347}       & \multicolumn{1}{r|}{{\underline{.523}}}         & \multicolumn{1}{r|}{.763}       & \multicolumn{1}{r|}{.446}       & \multicolumn{1}{r|}{.342}       & .517                                    \\ \hline
SW-Attn+BiLSTM                                                                                                    & \multicolumn{1}{r|}{.750}       & \multicolumn{1}{r|}{.446}       & \multicolumn{1}{r|}{.339}          & \multicolumn{1}{r|}{.511}               & \multicolumn{1}{r|}{.757}       & \multicolumn{1}{r|}{\textbf{.452}} & \multicolumn{1}{r|}{.332}       & \multicolumn{1}{r|}{.514}               & \multicolumn{1}{r|}{.763}       & \multicolumn{1}{r|}{.438}       & \multicolumn{1}{r|}{.345}       & .515                                    \\ \hline
SW-Attn+Encoder                                                                                                   & \multicolumn{1}{r|}{.765}       & \multicolumn{1}{r|}{.411}       & \multicolumn{1}{r|}{.319}          & \multicolumn{1}{r|}{.498}               & \multicolumn{1}{r|}{.767}       & \multicolumn{1}{r|}{.423}          & \multicolumn{1}{r|}{.327}       & \multicolumn{1}{r|}{.506}               & \multicolumn{1}{r|}{.763}       & \multicolumn{1}{r|}{.410}       & \multicolumn{1}{r|}{.343}       & .505                                    \\ \hline
\end{tabular}
\end{adjustbox}
\caption{Class-level F1 scores of the Sig-Networks toolkit models on \textbf{Anno-MI} for different History Lengths. \textbf{Best} and \underline{second best} scores are highlighted.}
\label{tab:anno-mi}
\end{table}

\begin{table}[htb!]
\begin{adjustbox}{max width=\linewidth}
\begin{tabular}{|l|crr|crr|crr|}
\hline
\multicolumn{1}{|c|}{\textbf{Model}}                                                                               & \multicolumn{3}{c|}{\textbf{No Switch}}                                                                                                               & \multicolumn{3}{c|}{\textbf{Switch}}                                                                                                                     & \multicolumn{3}{c|}{\textbf{Macro-avg}}                                                                                                               \\ \hline
BERT (focal)                                                                                                       & \multicolumn{3}{c|}{{\color[HTML]{000000} .724}}                                                                                                     & \multicolumn{3}{c|}{{\color[HTML]{000000} .454}}                                                                                                        & \multicolumn{3}{c|}{{\color[HTML]{000000} .589}}                                                                                                     \\ \hline
BERT (ce)                                                                                                          & \multicolumn{3}{c|}{{\color[HTML]{000000} .720}}                                                                                                     & \multicolumn{3}{c|}{{\color[HTML]{000000} .472}}                                                                                                        & \multicolumn{3}{c|}{{\color[HTML]{000000} .596}}                                                                                                     \\ \hline
FFN                                                                                                                & \multicolumn{3}{c|}{{\color[HTML]{000000} .704}}                                                                                                     & \multicolumn{3}{c|}{{\color[HTML]{000000} .457}}                                                                                                        & \multicolumn{3}{c|}{{\color[HTML]{000000} .581}}                                                                                                     \\ \hline
FFN History                                                                                                        & \multicolumn{3}{c|}{{\color[HTML]{000000} .727}}                                                                                                     & \multicolumn{3}{c|}{{\color[HTML]{000000} .523}}                                                                                                        & \multicolumn{3}{c|}{{\color[HTML]{000000} .625}}                                                                                                     \\ \hline
BiLSTM (w=5)                                                                                                       & \multicolumn{3}{c|}{{\color[HTML]{000000} .730}}                                                                                                     & \multicolumn{3}{c|}{{\color[HTML]{000000} .545}}                                                                                                        & \multicolumn{3}{c|}{{\color[HTML]{000000} .637}}                                                                                                     \\ \hline
SWNU (w=5)                                                                                                         & \multicolumn{3}{c|}{{\color[HTML]{000000} \textbf{.761}}}                                                                                            & \multicolumn{3}{c|}{{\color[HTML]{000000} .580}}                                                                                                        & \multicolumn{3}{c|}{{\color[HTML]{000000} .670}}                                                                                                     \\ \hline
SW-Attn (w=5)                                                                                                      & \multicolumn{3}{c|}{{\color[HTML]{000000} \textbf{.761}}}                                                                                            & \multicolumn{3}{c|}{{\color[HTML]{000000} .574}}                                                                                                        & \multicolumn{3}{c|}{{\color[HTML]{000000} .667}}                                                                                                     \\ \hline
\multicolumn{1}{|c|}{}                                                                                             & \multicolumn{3}{c|}{\textbf{11 (n=3)}}                                                                                                                & \multicolumn{3}{c|}{\textbf{20 (n=6)}}                                                                                                                   & \multicolumn{3}{c|}{\textbf{35 (n=11)}}                                                                                                               \\ \cline{2-10} 
\multicolumn{1}{|c|}{\multirow{-2}{*}{\textbf{\begin{tabular}[c]{@{}c@{}}History Length \\ (units)\end{tabular}}}} & \multicolumn{1}{c|}{\textbf{No Switch}}                 & \multicolumn{1}{c|}{\textbf{Switch}}              & \multicolumn{1}{c|}{\textbf{Macro-avg}} & \multicolumn{1}{c|}{\textbf{No Switch}}           & \multicolumn{1}{c|}{\textbf{Switch}}                       & \multicolumn{1}{c|}{\textbf{Macro-avg}} & \multicolumn{1}{c|}{\textbf{No Switch}}           & \multicolumn{1}{c|}{\textbf{Switch}}                    & \multicolumn{1}{c|}{\textbf{Macro-avg}} \\ \hline
BiLSTM                                                                                                             & \multicolumn{1}{r|}{{\color[HTML]{000000} .748}}       & \multicolumn{1}{r|}{{\color[HTML]{000000} .566}} & {\color[HTML]{000000} .657}            & \multicolumn{1}{r|}{{\color[HTML]{000000} .740}} & \multicolumn{1}{r|}{{\color[HTML]{000000} .555}}          & {\color[HTML]{000000} .648}            & \multicolumn{1}{r|}{{\color[HTML]{000000} .748}} & \multicolumn{1}{r|}{{\color[HTML]{000000} .548}}       & {\color[HTML]{000000} .648}            \\ \hline
SWNU                                                                                                               & \multicolumn{1}{r|}{{\color[HTML]{000000} .759}}       & \multicolumn{1}{r|}{{\color[HTML]{000000} .584}} & {\color[HTML]{000000} .671}            & \multicolumn{1}{r|}{{\color[HTML]{000000} .736}} & \multicolumn{1}{r|}{{\color[HTML]{000000} .571}}          & {\color[HTML]{000000} .654}            & \multicolumn{1}{r|}{{\color[HTML]{000000} .759}} & \multicolumn{1}{r|}{{\color[HTML]{000000} {\underline{.587}}}} & {\color[HTML]{000000} {\underline{.673}}}      \\ \hline
SW-Attn                                                                                                            & \multicolumn{1}{r|}{{\color[HTML]{000000} .745}}       & \multicolumn{1}{r|}{{\color[HTML]{000000} .573}} & {\color[HTML]{000000} .659}            & \multicolumn{1}{r|}{{\color[HTML]{000000} .747}} & \multicolumn{1}{r|}{{\color[HTML]{000000} .583}}          & {\color[HTML]{000000} .665}            & \multicolumn{1}{r|}{{\color[HTML]{000000} .743}} & \multicolumn{1}{r|}{{\color[HTML]{000000} .579}}       & {\color[HTML]{000000} .661}            \\ \hline
Seq-Sig-Net                                                                                                        & \multicolumn{1}{r|}{{\color[HTML]{000000} {\underline{.760}}}} & \multicolumn{1}{r|}{{\color[HTML]{000000} .584}} & {\color[HTML]{000000} .672}            & \multicolumn{1}{r|}{{\color[HTML]{000000} .754}} & \multicolumn{1}{r|}{{\color[HTML]{000000} \textbf{.602}}} & {\color[HTML]{000000} \textbf{.678}}   & \multicolumn{1}{r|}{{\color[HTML]{000000} .748}} & \multicolumn{1}{r|}{{\color[HTML]{000000} .559}}       & {\color[HTML]{000000} .654}            \\ \hline
SW-Attn+BiLSTM                                                                                                     & \multicolumn{1}{r|}{{\color[HTML]{000000} .742}}       & \multicolumn{1}{r|}{{\color[HTML]{000000} .584}} & {\color[HTML]{000000} .663}            & \multicolumn{1}{r|}{{\color[HTML]{000000} .741}} & \multicolumn{1}{r|}{{\color[HTML]{000000} .573}}          & {\color[HTML]{000000} .657}            & \multicolumn{1}{r|}{{\color[HTML]{000000} .750}} & \multicolumn{1}{r|}{{\color[HTML]{000000} .570}}       & {\color[HTML]{000000} .660}            \\ \hline
SW-Attn+Encoder                                                                                                    & \multicolumn{1}{r|}{{\color[HTML]{000000} .746}}       & \multicolumn{1}{r|}{{\color[HTML]{000000} .581}} & {\color[HTML]{000000} .664}            & \multicolumn{1}{r|}{{\color[HTML]{000000} .742}} & \multicolumn{1}{r|}{{\color[HTML]{000000} .572}}          & {\color[HTML]{000000} .657}            & \multicolumn{1}{r|}{{\color[HTML]{000000} .756}} & \multicolumn{1}{r|}{{\color[HTML]{000000} .569}}       & {\color[HTML]{000000} .662}            \\ \hline
\end{tabular}
\end{adjustbox}
\caption{Class-level F1 scores of the Sig-Networks toolkit models on \textbf{Longitudinal Rumour Stance} for different History Lengths. \textbf{Best} and \underline{second best} scores are highlighted.}
\label{tab:lrs}
\end{table}

\begin{table}[htb!]
\begin{adjustbox}{max width=\linewidth}
\begin{tabular}{|l|crrcrrcrrcrr|}
\hline
\multicolumn{1}{|c|}{\textbf{Model}}                                                                              & \multicolumn{3}{c|}{\textbf{IS}}                                                                                & \multicolumn{3}{c|}{\textbf{IE}}                                                                                  & \multicolumn{3}{c|}{\textbf{O}}                                                                                  & \multicolumn{3}{c|}{\textbf{Macro-avg}}                                                                            \\ \hline
BERT (focal)                                                                                                      & \multicolumn{3}{c|}{.283}                                                                                           & \multicolumn{3}{c|}{.439}                                                                                             & \multicolumn{3}{c|}{.871}                                                                                            & \multicolumn{3}{c|}{.531}                                                                                            \\ \hline
BERT (ce)                                                                                                         & \multicolumn{3}{c|}{.229}                                                                                           & \multicolumn{3}{c|}{.431}                                                                                             & \multicolumn{3}{c|}{\textbf{.903}}                                                                                   & \multicolumn{3}{c|}{.521}                                                                                            \\ \hline
FFN                                                                                                               & \multicolumn{3}{c|}{.281}                                                                                           & \multicolumn{3}{c|}{.432}                                                                                             & \multicolumn{3}{c|}{.890}                                                                                            & \multicolumn{3}{c|}{.534}                                                                                            \\ \hline
FFN History                                                                                                       & \multicolumn{3}{c|}{.280}                                                                                           & \multicolumn{3}{c|}{.454}                                                                                             & \multicolumn{3}{c|}{.877}                                                                                            & \multicolumn{3}{c|}{.537}                                                                                            \\ \hline
BiLSTM (w=5)                                                                                                      & \multicolumn{3}{c|}{.260}                                                                                           & \multicolumn{3}{c|}{.479}                                                                                             & \multicolumn{3}{c|}{.892}                                                                                            & \multicolumn{3}{c|}{.544}                                                                                            \\ \hline
SWNU (w=5)                                                                                                        & \multicolumn{3}{c|}{.301}                                                                                           & \multicolumn{3}{c|}{{\underline{.494}}}                                                                                       & \multicolumn{3}{c|}{{\underline{.894}}}                                                                                      & \multicolumn{3}{c|}{\textbf{.563}}                                                                                   \\ \hline
SW-Attn (w=5)                                                                                                     & \multicolumn{3}{c|}{.300}                                                                                           & \multicolumn{3}{c|}{.480}                                                                                             & \multicolumn{3}{c|}{.887}                                                                                            & \multicolumn{3}{c|}{.556}                                                                                            \\ \hline
\multicolumn{1}{|c|}{\multirow{2}{*}{\textbf{\begin{tabular}[c]{@{}c@{}}History Length \\ (units)\end{tabular}}}} & \multicolumn{4}{c|}{\textbf{11 (n=3)}}                                                                                                                        & \multicolumn{4}{c|}{\textbf{20 (n=6)}}                                                                                                                        & \multicolumn{4}{c|}{\textbf{35 (n=11)}}                                                                                                                       \\ \cline{2-13} 
\multicolumn{1}{|c|}{}                                                                                            & \multicolumn{1}{c|}{\textbf{IS}} & \multicolumn{1}{c|}{\textbf{IE}} & \multicolumn{1}{c|}{\textbf{O}} & \multicolumn{1}{c|}{\textbf{Macro-avg}} & \multicolumn{1}{c|}{\textbf{IS}} & \multicolumn{1}{c|}{\textbf{IS}} & \multicolumn{1}{c|}{\textbf{O}} & \multicolumn{1}{c|}{\textbf{Macro-avg}} & \multicolumn{1}{c|}{\textbf{IS}} & \multicolumn{1}{c|}{\textbf{IE}} & \multicolumn{1}{c|}{\textbf{O}} & \multicolumn{1}{c|}{\textbf{Macro-avg}} \\ \hline
BiLSTM                                                                                                            & \multicolumn{1}{r|}{.252}            & \multicolumn{1}{r|}{.478}            & \multicolumn{1}{r|}{.887}           & \multicolumn{1}{r|}{.539}             & \multicolumn{1}{r|}{.244}            & \multicolumn{1}{r|}{.470}            & \multicolumn{1}{r|}{.887}           & \multicolumn{1}{r|}{.533}             & \multicolumn{1}{r|}{.225}            & \multicolumn{1}{r|}{.460}            & \multicolumn{1}{r|}{.891}           & .525                                  \\ \hline
SWNU                                                                                                              & \multicolumn{1}{r|}{.292}            & \multicolumn{1}{r|}{.471}            & \multicolumn{1}{r|}{.887}           & \multicolumn{1}{r|}{.550}             & \multicolumn{1}{r|}{.275}            & \multicolumn{1}{r|}{.448}            & \multicolumn{1}{r|}{.888}           & \multicolumn{1}{r|}{.537}             & \multicolumn{1}{r|}{.270}            & \multicolumn{1}{r|}{.457}            & \multicolumn{1}{r|}{.889}           & .539                                  \\ \hline
SW-Attn                                                                                                           & \multicolumn{1}{r|}{.286}            & \multicolumn{1}{r|}{.471}            & \multicolumn{1}{r|}{.884}           & \multicolumn{1}{r|}{.547}             & \multicolumn{1}{r|}{.286}            & \multicolumn{1}{r|}{.453}            & \multicolumn{1}{r|}{.883}           & \multicolumn{1}{r|}{.541}             & \multicolumn{1}{r|}{.289}            & \multicolumn{1}{r|}{.452}            & \multicolumn{1}{r|}{.876}           & .539                                  \\ \hline
Seq-Sig-Net                                                                                                       & \multicolumn{1}{r|}{.301}            & \multicolumn{1}{r|}{\textbf{.495}}   & \multicolumn{1}{r|}{.893}           & \multicolumn{1}{r|}{\textbf{.563}}    & \multicolumn{1}{r|}{\textbf{.304}}   & \multicolumn{1}{r|}{.487}            & \multicolumn{1}{r|}{.891}           & \multicolumn{1}{r|}{{\underline{.561}}}       & \multicolumn{1}{r|}{{\underline{.303}}}      & \multicolumn{1}{r|}{.480}            & \multicolumn{1}{r|}{{\underline{.894}}}     & .559                                  \\ \hline
SW-Attn+BiLSTM                                                                                                    & \multicolumn{1}{r|}{.291}            & \multicolumn{1}{r|}{.483}            & \multicolumn{1}{r|}{.887}           & \multicolumn{1}{r|}{.554}             & \multicolumn{1}{r|}{.298}            & \multicolumn{1}{r|}{.483}            & \multicolumn{1}{r|}{.890}           & \multicolumn{1}{r|}{.557}             & \multicolumn{1}{r|}{.298}            & \multicolumn{1}{r|}{.467}            & \multicolumn{1}{r|}{.885}           & .550                                  \\ \hline
SW-Attn+Encoder                                                                                                   & \multicolumn{1}{r|}{.289}            & \multicolumn{1}{r|}{.477}            & \multicolumn{1}{r|}{.890}           & \multicolumn{1}{r|}{.552}             & \multicolumn{1}{r|}{.302}            & \multicolumn{1}{r|}{.463}            & \multicolumn{1}{r|}{.891}           & \multicolumn{1}{r|}{.552}             & \multicolumn{1}{r|}{.294}            & \multicolumn{1}{r|}{.452}            & \multicolumn{1}{r|}{.887}           & .545                                  \\ \hline
\end{tabular}
\end{adjustbox}
\caption{Class-level F1 scores of the Sig-Networks toolkit models on \textbf{TalkLife MoC} for different History Lengths. \textbf{Best} and \underline{second best} scores are highlighted.}
\label{tab:talklife}
\end{table}

\section{Time Feature Guidance} \label{app:guidance}

As mentioned in ~\S\ref{sec:datamodules} the toolkit allows for the automatic computation of the following time-derived features if a timestamp column is provided:

\begin{itemize}[nosep]
    \item \texttt{time\_encoding}: date as fraction of the year
    \item \texttt{time\_encoding\_minute}: time as fraction of minutes, ignoring the date
    \item \texttt{time\_diff}: time difference between consecutive data in the stream
    \item \texttt{timeline\_index}: index of the data point in the stream
   
\end{itemize}

The option to include user-processed time features is available. Optionally, the user can specify a standardisation method for each time feature from the list below:

\begin{itemize}[nosep]
    \item \texttt{None}: no transformation applied
    \item \texttt{z\_score}: transformation by subtracting the mean and dividing by the standard deviation of the data points
    \item \texttt{sum\_divide}: transformation by dividing by the sum of the data points
    \item \texttt{minmax}: transformation by subtracting the minimum of data points from the current data point and dividing by the differential of the maximum and minimum of the data points.
\end{itemize}

The above (normalised) features can be included as part of the path stream in the signature model (\textit{in-path}) and/or concatenated with the SBERT representation of the current data point in the input to the final FFN layers in the model (\textit{in-input}) . During the different task modeling we find particularly important the efficient incorporation of time features. Such decision is task-driven. 

For Anno-MI we include the \texttt{time\_encoding\_minute} and \texttt{timeline\_index} (without transformation) \textit{in-path}. For Longitudinal Rumour Stance we include \texttt{time\_encoding} normalised with \texttt{z\_score} and \texttt{timeline\_index} without normalisation both \textit{in-path} and \textit{in-input}. Finally for TalkLife MoC we use \texttt{time\_encoding} normalised with \texttt{z\_score} both \textit{in-path} and \textit{in-input}. Since TalkLife and Longitudinal Rumour Stance are social media datasets they can benefit from the use of \textit{in-input} features that model the temporal semantic component of linguistic representations. We expect \textit{in-input} features to be less beneficial for our specific dialogue task which is semantically stable with conversations being date-agnostic (but not time agnostic). At the same time in the dialogue task of Anno-MI, the use of both the \texttt{time\_encoding\_minute}, which ignores the date, and \texttt{timeline\_index} \textit{in-path}, allows for modeling both the temporal flow of the conversation and the position (index) of the utterance of interest in the dialogue. While Longitudinal Rumour Stance also benefits from using the \texttt{timeline\_index} which identifies the position of information with respect to the initial claim, the use of \texttt{time\_encoding} normalised with \texttt{z\_score} is more suitable here as it makes use of the date of the comment. In TalkLife only the latter is used, without any index features. Here, since relevant context for each post under consideration occurs in short history windows, the timeline position (index) is irrelevant. By presenting how different time features benefit each task together with the intuition behind the selection process, we encourage users to consider the temporal characteristics of their task in-hand for efficient time feature selection.

\section{Package Environment}

The experiments ran in a Python 3.8.17 environment with the key following libraries: \texttt{sig-networks} (0.2.0), \texttt{nlpsig} (0.2.2), \texttt{torch} (1.9.0), \texttt{signatory} (1.2.6.1.9.0), \texttt{sentence-transformers} (2.2.2), \texttt{transformers} (4.30.2), \texttt{accelerate} (0.20.1), \texttt{evaluate} (0.4.0), \texttt{datasets} (2.14.2), \texttt{pandas} (1.5.3), \texttt{numpy} (1.24.4),  \texttt{scikit-learn} (1.3.0), \texttt{umap} (0.5.3).

\section{Path Signature Libraries} \label{app:siglibraries}

\begin{table}[htb!]
\begin{adjustbox}{max width=\linewidth,center}
\begin{tabular}{|l|l|}
\hline
\multicolumn{1}{|c|}{\textbf{Library}} & \multicolumn{1}{c|}{\textbf{Link}}          \\ \hline
\texttt{roughpy}                                & \url{https://github.com/datasig-ac-uk/RoughPy}    \\ \hline
\texttt{esig}                                   & \url{https://github.com/datasig-ac-uk/esig}       \\ \hline
\texttt{iisignature}                            & \url{https://github.com/bottler/iisignature}      \\ \hline
\texttt{signatory}                              & \url{https://github.com/patrick-kidger/signatory} \\ \hline
\texttt{signax}                                 & \url{https://github.com/Anh-Tong/signax}          \\ \hline
\end{tabular}
\end{adjustbox}
\end{table}

\section{Infrastructure}

The experiments with the Anno-MI and Longitudinal Stance datasets were ran on the \emph{Baskerville}, a GPU Tier2 cluster developed and maintained by the University of Birmingham in a collaboration with a number of partners including The Alan Turing Institute. Baskerville provided us access with Nvidia A100 GPUs (40GB and 80GB variants).

The experiments with the TalkLife dataset were ran on \emph{Sanctus}, a Queen Mary University of London maintained server, with a x86\_64 processor, 80 CPUs, 384 GB of RAM and 3 Nvidia A30 GPUs.

\section{Using the model modules} \label{app:model_modules}

As noted in \S\ref{sec:system:model}, we provide PyTorch modules for each of components of our Sig-Network models to encourage novel integration into other systems. For example, the key building blocks in each of our models are the Signature Window units, SWNU \citet{tseriotou2023sequential} and SW-Attn, as discussed in \S\ref{sec:method:sw_models}. These can be easily accessed in the toolkit with a few lines of Python code.

For example, in code listings \ref{lst:swnu} and \ref{lst:swmhau} we can simply load in the SWNU and SW-Attn units and initialise an instance of the module in a few lines. For initialising SWNU in listing \ref{lst:swnu}, we define several arguments: the input channels of our stream, \texttt{input\_channels}=$10$, the number of output channels after the convolution-1d layer, \texttt{output\_channels=5}, whether to take the log-signature or standard signature transformation, \texttt{log\_signature=False}, the signature depth, \texttt{sig\_depth=3}, the dimension of the LSTM hidden state(s), \texttt{hidden\_dim=5}, the pooling strategy to obtain a final stream representation, \texttt{pooling="signature"}, to not chronologically reverse the order of the stream, \texttt{reverse\_path=False}, to use a BiLSTM, \texttt{BiLSTM=True}, to use a convolution-1d layer, \texttt{augmentation\_type="Conv1d"}. The alternative option for \texttt{augmentation\_type} is to have \texttt{augmentation\_type="signatory"} which will use the \texttt{signatory.Augment} PyTorch module to use a larger convolution neural network (CNN) for which you can specify the hidden dimensions to in the \texttt{hidden\_dim\_aug} argument which is set to \texttt{None} in this example. Note that some of these arguments have default values, but we present them all here for more clarity.

\begin{lstlisting}[language=Python, caption=Example initialisation of Signature Window Network Unit object, label={lst:swnu}]
from sig_networks.swnu import SWNU

# initialise a SWNU object
swnu = SWNU(
    input_channels=10,
    output_channels=5,
    log_signature=False,
    sig_depth=3,
    hidden_dim=5,
    pooling="signature",
    reverse_path=False,
    BiLSTM=True,
    augmentation_type="Conv1d",
)
\end{lstlisting}

The SW-Attn unit, called \texttt{SWMHAU} in the library, shares many of the same arguments as expected but since we are using Multihead-Attention (MHA) in place of a (Bi)LSTM, we specify the number of attention heads through the \texttt{num\_heads} argument and specify how many stacks of these layers through the \texttt{num\_layers} argument. We can also specify the dropout to use in the MHA layer here too.

\begin{lstlisting}[language=Python, caption=Example initialisation of SW-Attention unit object, label={lst:swmhau}]
from sig_networks.swmhau import SWMHAU

# initialise a SWMHAU object
swmhau = SWMHAU(
    input_channels=10,
    output_channels=5,
    log_signature=False,
    sig_depth=3,
    num_heads=5,
    num_layers=1,
    dropout_rate=0.1,
    pooling="signature",
    reverse_path=False,
    augmentation_type="Conv1d",
)
\end{lstlisting}

Note that there are variants of these PyTorch modules which do not include the convolution 1d or CNN to project down the stream to a lower dimension before taking expanding window signatures, namely \texttt{sig\_networks.SWLSTM} and \texttt{sig\_networks.SWMHA}.

Once these objects have been created, they can simply be called to apply a forward pass of the units, see for example listing \ref{lst:unit_call}. These units receive as input a three-dimensional tensor of the batched streams and the resulting output is a two-dimensional tensor of batches of the fixed-length feature representations of the streams.

\begin{lstlisting}[language=Python, caption=Example forward pass of SWNU and SWMHAU objects, label={lst:unit_call}]
    import torch
    
    # create a three-dimensional tensor of 100 batched streams, each with history length w and 10 channels
    streams = torch.randn(100, 20, 10)
    
    # pass the streams through the SWNU
    # swnu_features and swmhau_features are two-dimensional tensors of shape [batch, signature_channels]
    swnu_features = swnu(streams)
    swmhau_features = swmhau(streams)
\end{lstlisting}

For full examples on how these PyTorch modules can be fitted into larger PyTorch networks, please refer to the source code for the Sig-Network family models in the library on GitHub\footnote{https://github.com/ttseriotou/sig-networks/tree/main/src/sig\_networks}.